\documentclass[letterpaper]{article} % DO NOT CHANGE THIS
\usepackage{aaai25}  % DO NOT CHANGE THIS
\usepackage{times}  % DO NOT CHANGE THIS
\usepackage{helvet}  % DO NOT CHANGE THIS
\usepackage{courier}  % DO NOT CHANGE THIS
\usepackage[hyphens]{url}  % DO NOT CHANGE THIS
\usepackage{graphicx} % DO NOT CHANGE THIS
\usepackage{natbib}  % DO NOT CHANGE THIS AND DO NOT ADD ANY OPTIONS TO IT
\usepackage{csquotes}
\usepackage{caption} % DO NOT CHANGE THIS AND DO NOT ADD ANY OPTIONS TO IT
\usepackage{algorithm}
\usepackage{algorithmic}

\usepackage{newfloat}
\usepackage{listings}
\usepackage{url}
\DeclareCaptionStyle{ruled}{labelfont=normalfont,labelsep=colon,strut=off} % DO NOT CHANGE THIS
\floatstyle{ruled}
\newfloat{listing}{tb}{lst}{}
\floatname{listing}{Listing}
\newcommand{\model}{\epsilon_\theta}
\newcommand{\conditioner}{\tau_\theta}

\newcommand{\cond}{y}
\usepackage{color, colortbl}
\usepackage{amsfonts}
\usepackage{amsmath}
\definecolor{Gray}{gray}{0.9}
\definecolor{Red}{RGB}{230, 57, 70}
\newcommand{\NewPara}[1]{\vspace{.05in}\noindent\textbf{#1}}
\usepackage{multirow}
\usepackage{pifont}
\usepackage{booktabs}
\usepackage{subcaption}
\title{Tri-Ergon: Fine-grained Video-to-Audio Generation with\\Multi-modal Conditions and LUFS Control}
\author {
    Bingliang Li\textsuperscript{\rm 1,2}\equalcontrib,
    Fengyu Yang\textsuperscript{\rm 2}\equalcontrib,
    Yuxin Mao\textsuperscript{\rm 3},
    Qingwen Ye\textsuperscript{\rm 1},
    Hongkai Chen\textsuperscript{\rm 1}\thanks{Corresponding author.},
    Yiran Zhong\textsuperscript{\rm 4 †}
}

\affiliations {
    \textsuperscript{\rm 1}vivo Mobile Communication Co., Ltd\\
    \textsuperscript{\rm 2}The Chinese University of Hong Kong, Shenzhen\\
    \textsuperscript{\rm 3}Northwestern Polytechnical University\\
    \textsuperscript{\rm 4}OpenNLPLab\\

    {\tt\small \{bingliangli,fengyuyang1\}@link.cuhk.edu.cn, maoyuxin@mail.nwpu.edu.cn, qingwen.ye@vivo.com, allenhkchen@gmail.com,  zhongyiran@gmail.com}
}
\usepackage{bibentry}
\begin{document}

\maketitle

\begin{abstract}
Video-to-audio (V2A) generation utilizes visual-only video features to produce realistic sounds that correspond to the scene. However, current V2A models often lack fine-grained control over the generated audio, especially in terms of loudness variation and the incorporation of multi-modal conditions. To overcome these limitations, we introduce Tri-Ergon, a diffusion-based V2A model that incorporates textual, auditory, and pixel-level visual prompts to enable detailed and semantically rich audio synthesis. Additionally, we introduce Loudness Units relative to Full Scale (LUFS) embedding, which allows for precise manual control of the loudness changes over time for individual audio channels, enabling our model to effectively address the intricate correlation of video and audio in real-world Foley workflows. Tri-Ergon is capable of creating 44.1 kHz high-fidelity stereo audio clips of varying lengths up to 60 seconds, which significantly outperforms existing state-of-the-art V2A methods that typically generate mono audio for a fixed duration.
\end{abstract}

\begin{links}
    \link{Project website}{https://tri-ergon.github.io/Tri-Ergon/}
    % \link{Datasets}{https://aaai.org/example/datasets}
    % \link{Extended version}{https://aaai.org/example/extended-version}
\end{links}

\section{Introduction}
{When we experience visual events, we naturally expect to hear the corresponding sounds, enhancing our understanding and perception of the world. However, current video generation models mainly focus on creating visual content from text, often neglecting the integration of audio. The video-to-audio (V2A) task aims to fill this gap by generating semantically relevant and temporally synchronized audio from video frames. Due to its potential applications in movie automatic dubbing, video content production, game production, and other fields, the V2A task has attracted growing interest. As video generation models undergo continual refinement, the generation of fine-grained, high-fidelity, synchronized audio that aligns with prescribed loudness standards has emerged as a pivotal focal point for research.
}

Recent innovations in V2A generation have demonstrated considerable progress; however, current state-of-the-art methods still face limitations that hinder their practical application in industrial settings. Three primary issues have been identified in existing methods:
i) The generated audio content may not be closely aligned with the corresponding visual or textual inputs due to single/double modality inputs.
ii) Lack of precise control over audio attributes to effectively synchronize the audio and visual contents.
iii) The absence of high fidelity, stereo, long-duration, and open-domain audio generation capability.
Overcoming these challenges is essential for improving the effectiveness of V2A models in accurately generating audio alongside visual content.

In this paper, we posit that these challenges can be effectively mitigated through the utilization of \emph{fine-grained audio control and generation}. Our concept of fine-grained encompasses two key aspects: i) the application of multi-modal conditions for audio generation, and ii) the implementation of a meticulous audio generation process that allows for exact control over loudness, audio quality, and duration. The former enables the model to leverage a broader range of information for audio generation, resulting in more coherent audio output. For instance, text information can provide semantic context for the audio, which is particularly beneficial for off-screen audio generation. Audio input serves as a style reference for the audio generation process. The latter aspect of fine-grained control ensures that the generated audio is closely aligned with the visual input and possesses enhanced quality.

\begin{figure*}
    \centering
    \includegraphics[width=0.95\linewidth]{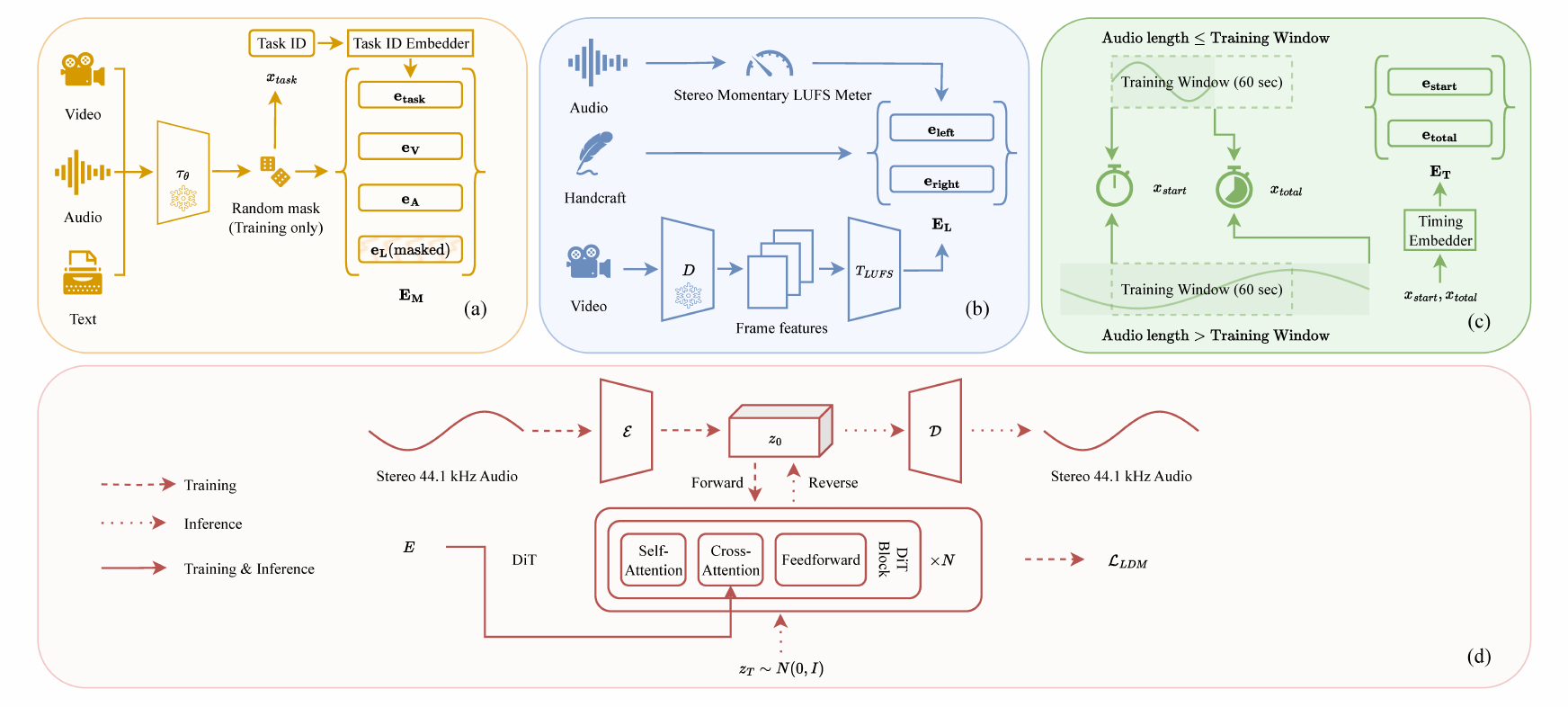}
    % \vspace{-8mm}
    \caption{\textbf{Tri-Ergon Overview.} (a) Unbalanced multi-modal prompting module, which takes any combination of prompts. (b) Stereo loudness control with LUFS. (c) Timing embeddings for variable audio length. (d) The DiT backbone.}
    \label{fig:enter-label}
\end{figure*}

In light of the aforementioned observation, we propose a novel framework called Tri-Ergon\footnote{In memory of the sound-on-film system developed around 1919~\cite{lipton2021cinema} by three German inventors. Tri-Ergon means “the work of three”, derived from the Greek word $\tau\rho\acute{\iota}\alpha$, which corresponds to the three modalities in our work.} to address the above limitations, as depicted in Figure~\ref{fig:enter-label}. Tri-Ergon introduces three multi-modal conditioning and precise loudness control mechanisms. Specifically, the multi-modal conditioning utilizes textual, auditory, visual prompts to steer the audio generation process, enabling the creation of audio tracks that are semantically rich and contextually appropriate. An unbalanced multi-modal prompting module is introduced to enable Tri-Ergon to perform audio generation from any modality with aligned, semantic-rich features~\cite{zhu2023languagebind} from video, text, and audio.

Tri-Ergon incorporates Loudness Units relative to Full Scale (LUFS) for precise loudness control mechanisms. LUFS provides a standardized metric closely aligned with human auditory perception, allowing fine-grained control over audio loudness throughout its entire duration. The momentary LUFS can be derived from a reference audio, crafted by an audio engineer, or predicted using our LUFS prediction module, which utilizes DINO-V2 to extract frame features from video and a transformer encoder to obtain the predicted LUFS.
By integrating LUFS into the diffusion process, we enable precise adjustments to loudness variations over time, crucial for maintaining audio consistency and quality, especially in complex Foley workflows where synchronization between audio and visual elements is essential.

We have created the first comprehensive V2A dataset for training Tri-Ergon, featuring high-fidelity, long-duration, open-vocabulary multi-modal labeling, named MM-V2A. Through training on this dataset, Tri-Ergon can generate top-quality stereo audio at a sampling rate of 44.1 kHz. It is capable of producing audio clips of any length, with a maximum duration of 60 seconds. This represents a significant advancement over current V2A methods, which are generally limited to generating mono audio for a fixed duration.

The main contributions can be summarized as follows:
\begin{itemize}
    \item We propose Tri-Ergon, a novel framework for video-to-audio (V2A) generation that introduces fine-grained multi-modal conditioning to generate semantically rich and contextually aligned audio tracks.
        
    \item We incorporate Loudness Units relative to Full Scale (LUFS) into the diffusion process, enabling precise control over loudness variations throughout the audio generation process.

    \item We introduce MM-V2A, a comprehensive V2A dataset with high-fidelity, long-duration, open-vocabulary multi-modal labeling for training our proposed Tri-Ergon.
    
    \item Our framework outperforms existing state-of-the-art methods, as demonstrated by both qualitative and quantitative results.
\end{itemize}

\section{Related Work}
\label{related_work}

\NewPara{Video-to-Audio Generation.}
Video-to-audio (V2A) generation focuses on synthesizing audio that aligns with the visual content of a video. SpecVQGAN~\cite{iashin2021taming} uses a Transformer-based autoregressive model with ResNet50 or RGB+Flow as the visual backbone to predict spectrograms from visual features, thereby generating audio. Im2Wav~\cite{sheffer2023hear} employs a dual-transformer model conditioned on CLIP features to produce sound directly from visual data. Difffoley~\cite{luo2024diff} enhances the coherence of generated audio by using contrastive audio-visual pretraining to align audio and visual features. Some approaches also incorporate multi-modal and joint training encoders. V2A-Mapper~\cite{wang2024v2a} uses a lightweight mapper to convert CLIP embeddings of videos into CLAP~\cite{wu2023large} embeddings, conditioning the audio generation on visual input. Additionally, Xing et al.~\cite{xing2024seeing} leverage an ImageBind-based latent aligner for conditional guidance in audio generation.

\NewPara{Multi-modal representation.}
Multi-modal representation typically starts with vision and language pretraining. CLIP~\cite{radford2021learning} pioneered this approach by aligning images and texts using a large-scale dataset of 400 million samples, effectively bridging the gap between these two domains. Building on CLIP's success, various models have extended this alignment to other modalities. CLIP4Clip~\cite{luo2022clip4clip} aligns videos with text, and CLAP aligns audio with text. Furthermore, more than two modalities can be aligned simultaneously with some modifications. ImageBind~\cite{girdhar2023imagebind} broadens multi-modal alignment pretraining to include six different modalities. LanguageBind~\cite{zhu2023languagebind}, a language-based multi-modal pretraining approach, aligns all modalities with the language modality through contrastive learning, unifying them within a shared embedding space during the pretraining process.

\NewPara{High-Quality and Variable-Length Audio Generation.}
In terms of quality, AudioLDM2~\cite{liu2024audioldm} demonstrates the ability to generate 48kHz mono audio, while Levy et al.~\cite{levy2023controllable} achieve the generation of 44.1kHz stereo music. Regarding the duration of generated audio, previous non-autoregressive models were constrained to producing music of up to 20 seconds~\cite{parker2024stemgen}. End-to-end and latent diffusion models extended this limit to 30 seconds~\cite{levy2023controllable, huang2023noise2music}. Stable Audio~\cite{evans2024fast} further advances the state-of-the-art by generating stereo signals up to 95 seconds in length. Our proposed method, Tri-Ergon, utilizes Diffusion Transformers to deliver high-fidelity stereo audio at 44.1kHz, with variable-length outputs of up to 60 seconds.
\section{Methodology}
\label{method}

\subsection{Preliminary}\label{sec:ldm}
Our framework is based on the latent diffusion model, which follows the standard formulation outlined in DDPM, and comprises a forward diffusion process and a backward reverse denoising process. 
Initially, a data sample $\mathbf{x} \sim p(\mathbf{x})$ undergoes processing by an autoencoder, consisting of an encoder $\mathcal{E}$ and a decoder $\mathcal{D}$. 
The autoencoder projects $\mathbf{x}$ into a latent variable $\mathbf{z}$ via $\mathbf{z}=\mathcal{E}(\mathbf{x})$. Subsequently, the diffusion and denoising process takes place within the latent space.

The denoised latent variable is recovered to the input space by $\hat{\mathbf{x}}=\mathcal{D}(\mathbf{\hat{z}_0})$.

Inspired by non-equilibrium thermodynamics, diffusion models~\cite{ho2020denoising} are a class of latent variable ($z_1, ..., z_T$) models of the form 
$p_{\theta}(z_{0}) = \int p_{\theta}(z_{0:T}) d z_{1:T}$, 
where the latent variables are of the same dimensionality as the input data $z_0$. 
The joint distribution $p_{\theta}(z_{0:T})$ is also called the \textit{reverse process}:
\begin{equation}
    p_{\theta}(z_{0:T}) = p_{\theta}(z_T)\prod_{t=1}^{T} p_{\theta}(z_{t-1}|z_{t}),
\end{equation}
\begin{equation} \label{eq: reverse}
    p_{\theta}(z_{t-1}|z_{t}) = \mathcal{N}(z_{t-1};\mu_\theta(z_t,t),\Sigma_\theta(z_t,t)).
\end{equation}
Here, $\mu_\theta$ and $\Sigma_\theta$ are determined through a denoiser network $\epsilon_{\theta}(z_t,t)$, typically structured as a UNet or a transformer~\cite{ronneberger_unet_2015, peebles2023scalable, mao2023contrastive}.

The approximate posterior $q(z_{1:T}|z_{0})$ is called the \textit{forward process}, which is fixed to a Markov chain that gradually adds noise according to a predefined noise scheduler $\beta_{1:T}$:
\begin{equation}
    q(z_{1:T}|z_{0}) = \prod_{t=1}^{T} q(z_{t}|z_{t-1}),
    \label{eq:z0_to_zt}
\end{equation}
\begin{equation} \label{eq: diffusion}
    q(z_{t}|z_{t-1}) = \mathcal{N}(z_{t};\sqrt{1-\beta_{t}}z_{t-1},\beta_{t}\mathbf{I}).
\end{equation}

The training is performed by minimizing a variational bound on the negative log-likelihood, and the final training objective for $\theta$ is a noise estimation loss with a conditional variable $\mathbf{E}$. It can be formulated as:
\begin{equation}
    \begin{aligned}
    \label{ddpm_loss_condition}
    \mathcal{L}_{\text{diffusion}}(\theta):=\mathbb{E}_{\mathbf{z}, \mathbf{c}, \epsilon\sim \mathcal{N}(0,\mathbf{I}), t}\left[\left\|\epsilon-\epsilon_\theta\left(\mathbf{z}_t, \mathbf{E}, t\right)\right\|^2\right].
    \end{aligned}
\end{equation}

\subsection{Conditioning}
The conditional variable $\mathbf{E}$, as highlighted in the diffusion loss function, plays a crucial role in guiding the generation.

\subsubsection{Unbalanced multi-modal prompting module.}
\label{unbalanced}

We introduce an unbalanced multi-modal prompting module as a conditioning mechanism, leveraging LanguageBind as the multi-modal encoder. We denote the LanguageBind encoder as $\conditioner$, which encodes the input condition $\cond$ (e.g., language, audio, and video) into an embedding $\conditioner(\cond) \in \mathbb{R}^{M \times d_\tau}$, represented as $\mathbf{e_L}$, $\mathbf{e_A}$, and $\mathbf{e_V}$, respectively.

To enhance the ability to generate audio based on various combinations of conditions, we propose a random masking mechanism. During training, the embeddings $\mathbf{e_L}$, $\mathbf{e_A}$, and $\mathbf{e_V}$ are randomly masked, leaving between 0 and 3 conditions active. Additionally, a task ID $x_{task}$ is assigned based on the combination of available conditions (e.g., $\langle \mathbf{e_L}, \text{mask}, \mathbf{e_V} \rangle$ is assigned the integer 5). This task ID is then encoded into a task embedding $\mathbf{e_{task}}$.

The final multi-modal feature representation $\mathbf{E_M} \in \mathbb{R}^{4M \times d_\tau}$ is constructed by stacking $\mathbf{e_{task}}$, $\mathbf{e_L}$, $\mathbf{e_A}$, and $\mathbf{e_V}$. This representation contains aligned, semantically rich information, enabling our model to generate audio based on prompts from up to three modalities.

\subsubsection{Stereo loudness control with LUFS.}
Loudness Units relative to Full Scale (LUFS) is a standardized metric for audio loudness that accounts for human perception and signal intensity, widely used for audio normalization in media. LUFS can be measured in three ways: momentary (400 ms window), short-term (3-second moving average), and integrated (mean of short-term LUFS). We utilize a customized momentary LUFS measurement with a window size of $\frac{1}{6}$ second.

For a given audio signal $x_a$ with duration $L_a$, the LUFS information is extracted separately for the left and right channels, resulting in $\mathbf{e'_{left}}$ and $\mathbf{e'_{right}}$, both with a length of $6L_a$. These values are clipped to the range of -70 (extremely quiet) to 0 (extremely loud) and normalized to a range of 0 to 1. The normalized values are then padded and interpolated to match the dimensions of $\mathbf{E_M}$ as $[\mathbf{e_{left}}, \mathbf{e_{right}}]$, forming the LUFS embedding $\mathbf{E_L} \in \mathbb{R}^{2M \times d_\tau}$.

We propose three methods for generating $\mathbf{E_L}$: (1) directly measuring a reference audio using a LUFS meter, (2) manually crafting $\mathbf{e'_{left}}$ and $\mathbf{e'_{right}}$ based on the video, and (3) utilizing our proposed LUFS prediction module, which can also be modified by the user for enhanced control.

The LUFS prediction module employs the DINO-V2 model~\cite{oquab2023dinov2} $D$ to extract image features from a video at 6 FPS, which are then processed by a transformer encoder $T_{LUFS}$ to predict the LUFS embedding $\mathbf{E_L}$. This module is trained using MSE loss and supports videos up to 60 seconds in length.

\subsubsection{Timing embeddings for variable audio length.}
The current V2A model often utilizes video encoders that can only process a fixed number of frames, and their audio VAE usually operates in the Mel spectrogram space. Due to these limitations, they can only generate audio with a limited and fixed duration, usually 10 seconds. To enable generating audio with variable length, we adopt the timing embeddings in~\cite{evans2024fast}.

When extracting a segment of audio from our training data, we determine two properties: the starting second of the segment, $x_{start}$, and the total duration of the original audio file, $x_{total}$. For instance, for a given audio, if we extract a 60-second segment from a 180-second audio file, starting at the 21.5-second mark, then $x_{start}$ would be 21.5, and $x_{total}$ would be 180. These values are subsequently converted into learned timing embeddings $\mathbf{E_T} \in \mathbb{R}^{2M\times d_\tau}$ and concatenated along the sequence dimension with $\mathbf{E_M}$ and $\mathbf{E_L}$ as $\mathbf{E}$ before being input into the DiT’s cross-attention layers.

During training, we randomly cut the audio of a video to perform augmentation. During inference, $x_{start}$ is always set to 0, and $x_{total}$ is set to the duration of the video. This enables the model to generate audio with variable length to exactly match the duration of the video.

\subsection{Multi-conditioned Latent Diffusion}
\label{diffusion_model}

\noindent\textbf{Latent autoencoder.}
We adopt the Variational Autoencoder (VAE) $\mathcal{E_\theta}$ in~\cite{evans2024stable} to achieve latent space encoding and decoding. Specifically, the encoding process aims to compress the raw high-fidelity stereo 44.1kHz audio $x_a$ into the lower-dimensional latent code
$\text{\footnotesize $z_0=\mathcal{E_\theta}(x_a) \in \mathbb{R}^{C\times\frac{L}{r}}$}$, where ${C}$ is the dimension of the encoded latent, ${L}$ is the length of the audio signal, ${r}$ is the downsampling factor, $\frac{r}{2 \times C}$ is the compression ratio.
We use a combination of multiple datasets to train the VAE, with specific details provided in the experiment section.

\noindent\textbf{Denoising network architecture.}
We utilize the Diffusion Transformer (DiT)~\cite{peebles2023scalable} as our denoising network $\model$. To effectively integrate conditioning into the diffusion process, cross-attention mechanisms are incorporated within the transformer's layers. These mechanisms allow the model to incorporate additional contextual information, such as text, visual, or auditory cues.

\noindent\textbf{Denoising with LUFS information.}
For video to audio generation, the objective is to generate synchronized, high-fidelity stereo audio $x_a$ given a multi-modal feature representation $\mathbf{E_M}$.

\noindent\textbf{Latent Diffusion Model Objective.}
In the context of multi-conditioned latent diffusion, the Latent Diffusion Model (LDM) aims to reverse the process of noise addition, which transforms the original data distribution into a standard Gaussian distribution over a series of timesteps. The goal of the LDM is to recover the original latent representation by minimizing the discrepancy between the actual noise added during the forward process and the model’s predicted noise.

Given the latent variable $z_t$ at timestep $t$, the model predicts the noise $\epsilon_\theta(z_t, t, \{\mathbf{E_M}, \mathbf{E_L}, \mathbf{E_T}\})$, where ${\mathbf{E_M}, \mathbf{E_L}, \mathbf{E_T}}$ represent the multi-modal feature embeddings that provide additional contextual information. The model is trained to minimize the denoising objective:
\begin{equation}
    % \footnotesize
    \mathcal{L}_{LDM} = \mathbb{E}_{z_0, t, \epsilon}\|\epsilon - \epsilon_\theta(z_t, t, \{\mathbf{E_M}, \mathbf{E_L}, \mathbf{E_T}\})\|_2^2 
\end{equation}

This objective function, based on $L_2$ loss (mean squared error), ensures that the model learns to effectively reverse the diffusion process, gradually removing noise from the latent code and reconstructing a clean representation.

\noindent\textbf{Reverse Process and Generation.}
During generation, the LDM starts with a latent code sampled from a standard Gaussian distribution and iteratively applies the learned denoising steps in reverse. Each step refines the latent code, guided by the multi-modal inputs, until a denoised latent representation is achieved. This refined latent code is then decoded by the VAE into the final high-fidelity output, such as stereo audio synchronized with video content. The integration of LUFS information and other contextual features ensures that the generated audio is both high-quality and consistent with the input modalities.
\section{Experiments}
\label{experiments}

\subsection{Experiment Settings}

\begin{figure*}[!ht]
    \centering
    \includegraphics[width=0.85\linewidth]{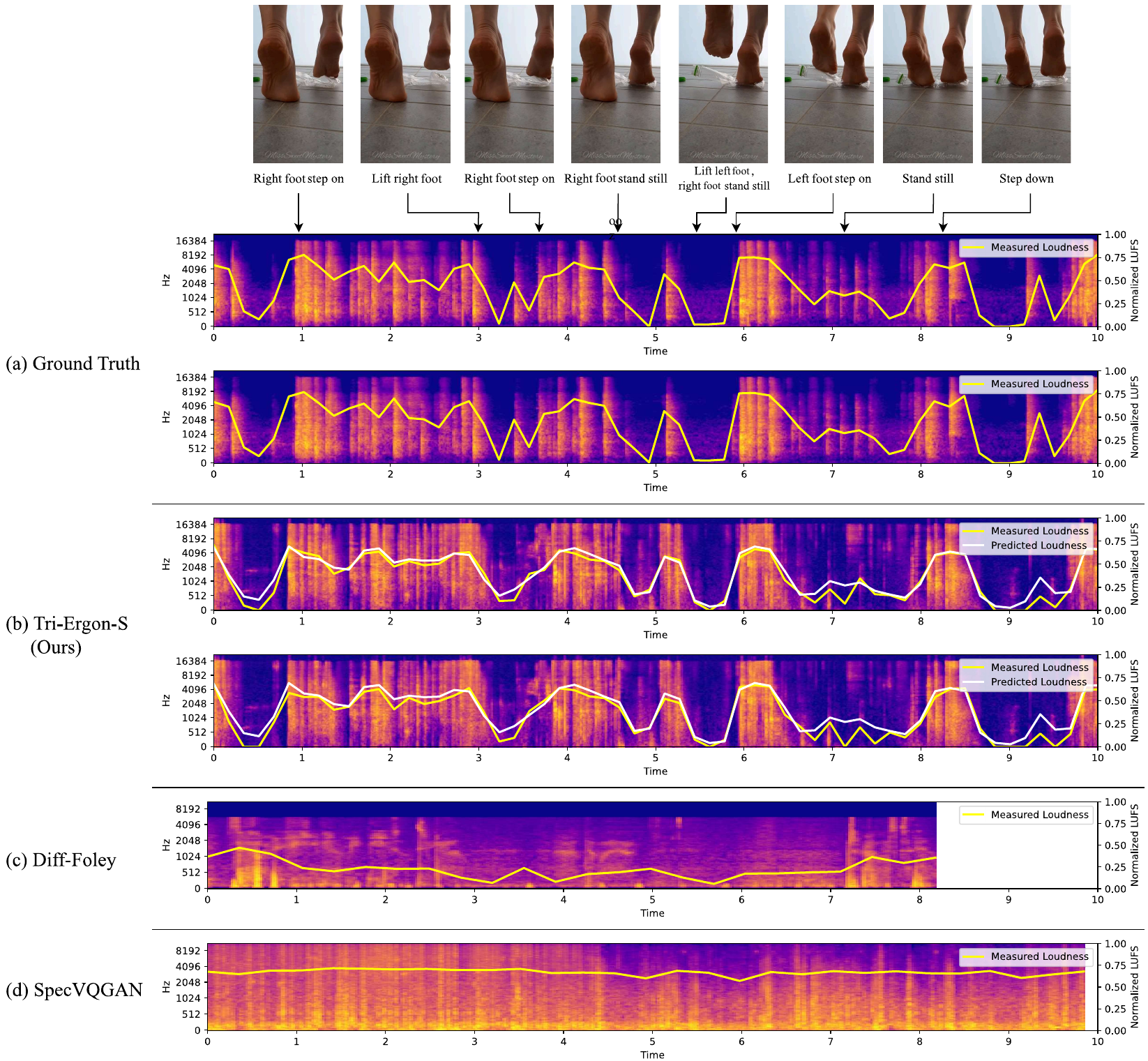}
    % \vspace{-3mm}
    \caption{Video-to-Audio generation results on VGGSound. 
    Given a video of a person alternately stepping on a plastic bottle with both feet, our LUFS prediction module can successfully detect the loudness variation and generate highly synchronized audio. Note that the predicted LUFS and the actual LUFS for the generated audio are close to each other, demonstrate the controllability Tri-Ergon provides.}
    \label{fig:demo_lufs}
\end{figure*}

\begin{table*}[h]
\centering
\small
\setlength{\tabcolsep}{0.7cm}
\begin{tabular}{@{}lcccc|c}
\toprule
Dataset                  & Video  & Text   & Audio  & Video Duration     & Unified Data \\ \midrule
VGGSound                 & 199K   & 199K    & 199K    & 10 sec             & 199K         \\
VALOR-32K                & 25.3K  & 25.3K  & 25.3K  & 10 sec             & 25.3K        \\
VIDAL-10M                & 69.7K  & 69.7K  & 69.7K  & $\textless$ 20 sec    & 69.7K        \\
BBC Sound Effect Library & -      & 31.1K  & 31.1K  & -                  & 31.1K        \\
FAVDBench                & 7.5K   & 7.5K   & 7.5K   & 5 sec $\sim$10 sec & 7.5K         \\
HD-VILA-100M             & 44.6K  & 44.6K  & 44.6K  & $\textless$ 60 sec   & 44.6K        \\
AudioSet                 & 128.1K & 128.1K & 128.1K & 10 sec             & 128.1K       \\ \midrule
MM-V2A                   & 474.2K & 505.3K & 505.3K & 5 sec $\sim$ 60 sec & 505.3K       \\ \bottomrule
\end{tabular}
\caption{Statistics for MM-V2A, which consolidates seven existing datasets, are presented in the table. The table numerically displays the count of attributes integrated from each source dataset.}
\label{tab:dataset_stat}
\end{table*}
\begin{table*}[t]
\centering
\small
\setlength{\tabcolsep}{0.53cm}
\begin{tabular}{@{}lccccc@{}}
\toprule
                                                        & Channels/sr  & Output length  & $\text{FD}_{openl3}$ $\downarrow$ & $\text{KL}_{passt}$ $\downarrow$ & AV-Align $\uparrow$ \\ \midrule
Training data (upper bound)                             & $2/44.1$kHz  & $10$ sec       & $16.87$                           & -                                & -                   \\
Autoencoded training data                               & $2/44.1$kHz  & $10$ sec       & $28.34$                           & -                                & -                   \\ \midrule
SpecVQGAN                                               & $1/22.05$kHz & $9.8$ sec      & $183.07$    & $3.12$                           & $0.204$             \\
Diff-Foley (Classifier-Free Guidance)                   & $1/16$kHz    & $8$ sec        & $249.15$                          & $3.07$                           & $0.177$             \\
Diff-Foley (Double Guidance)                            & $1/16$kHz    & $8$ sec        & $242.90$                          & $3.03$                           & $0.182$             \\
Tri-Ergon-S (Ours)                                      & $\mathbf{2/44.1}$kHz  & $10$ sec       & $144.57$                          & $2.63$                           & $0.229$             \\
Tri-Ergon-L (Ours)                                      & $\mathbf{2/44.1}$kHz  & \textbf{Up to $60$ sec} & $\mathbf{113.21}$               &  $\mathbf{1.82}$                  & $\mathbf{0.231}$    \\ \bottomrule
\end{tabular}
\caption{\footnotesize{Video-to-Audio generation results on VGGSound dataset. Our model achieve impressive results on both audio quality and temporal synchronization.} 
% \vspace{-3mm}
\label{tab:v2a_results}}
\end{table*}

\subsection{Model Configuration and Architecture Details} 
\subsubsection{VAE.}
We first train the audio VAE using automatic mixed precision for 1M steps with an effective batch size of 128 on 8 A100 GPUs. 
Similar to~\cite{evans2024fast}, we use the multi-resolution sum and difference STFT loss~\cite{steinmetz2021automatic} with A-weighting~\cite{fletcher1933loudness}, which is designed for stereo signals. We also utilize adversarial and feature matching losses with a modified multi-scale STFT EnCodec based discriminator~\cite{defossez2022high} that is adapted to process stereo audio. The losses are weighted as follows: $1.0$ for spectral losses, $0.1$ for adversarial losses, $5.0$ for feature matching loss, and $1\times 10^{-1}$ for KL loss. The total number of parameters for our VAE is 157M.

\subsubsection{Multi-conditioned Latent Diffusion.}
We define two model variants, Tri-Ergon-S and Tri-Ergon-L, based on their scale and number of parameters.
For the Tri-Ergon-S model, we employ a DiT architecture comprising 20 DiT blocks as the diffusion backbone, with 24 attention heads, an embedding dimension of 1536, and a total of 881M parameters. 
This model is trained for $7.2\times 10^5$ steps. 
Additionally, for Tri-Ergon-L, we utilize 24 DiT blocks while maintaining the same block configuration. It contains 1.1B parameters and is trained for $2.2\times 10^5$ steps. 
Both models are trained using the v-objective~\cite{salimans2022progressive}, with a cosine noise schedule and continuous denoising timesteps. The training process was conducted on 32 A100 GPUs, with an effective batch size of 256.
For the conditioning set $E_m$, we apply a 30\% probability of independently dropping the embeddings $\mathbf{e_L}$, $\mathbf{e_A}$, and $\mathbf{e_V}$. Additionally, a 10\% dropout rate is applied to the overall conditioning set $E$ to enable classifier-free guidance.

\subsubsection{LUFS prediction module.}
We train two separate LUFS prediction modules for Tri-Ergon-S and Tri-Ergon-L, referred to as $T_{LUFS-S}$ and $T_{LUFS-L}$, respectively.
The $T_{LUFS-S}$ module is based on a transformer encoder architecture with 12 layers, 8 attention heads, and an embedding dimension of 768. In contrast, $T_{LUFS-L}$ features a more complex architecture with 14 layers and 12 attention heads, while maintaining the same embedding dimension of 768.
$T_{LUFS-S}$ is trained on the VGGSound dataset for 10 epochs, whereas $T_{LUFS-L}$ is trained on our proposed dataset for 30 epochs. 
Both modules are trained on 8 A100 GPUs with a batch size of 128.

\subsubsection{Evaluation Metrics.} 
SpecVQGAN~\cite{iashin2021taming} relies on metrics such as Mean MKL and Melception-based FID for evaluating relevance and fidelity, using audio that is resampled to 22.05 kHz, mono, and fixed at 10 seconds. However, our work focuses on generating long-form full-band stereo signals, for which we utilize $FD_{openl3}$, $KL_{passt}$~\cite{evans2024fast}, and AV-Align~\cite{yariv2024diverse} as quantitative metrics. The $FD_{openl3}$ metric evaluates the similarity between generated and reference audio by projecting them into the Openl3 feature space and calculating the Fréchet Distance, making it suitable for assessing stereo signals up to 44.1 kHz. The $KL_{passt}$ metric leverages the PaSST model to compute the Kullback-Leibler (KL) divergence between the label probabilities of generated and reference audio, thereby assessing their semantic similarity by analyzing long-form audio segments. Additionally, AV-Align is utilized to measure the alignment between audio and visual modalities by detecting and comparing energy peaks in both.

\subsection{Datasets} 
The dataset used for training the diffusion model is systematically partitioned into two distinct segments: one dedicated to the training of the Tri-Ergon-S model and the other to the training of the Tri-Ergon-L model. The Tri-Ergon-S model is trained and evaluated exclusively on the VGG-Sound dataset~\cite{chen2020vggsound}, which consists of 200,000 video clips across 309 audio classes. Each 10-second clip features visual frames of the sound-producing object, paired with an audio track that corresponds to the visual content.

In contrast, the Tri-Ergon-L model is trained on a newly proposed and more comprehensive dataset, termed MM-V2A. This dataset is designed to extend beyond VGG-Sound by incorporating additional datasets with multi-modal annotations, including VALOR-32K~\cite{chen2024vast}, VIDAL-10M~\cite{zhu2023languagebind}, the BBC Sound Effect Library\footnote{https://sound-effects.bbcrewind.co.uk}, FAVDBench~\cite{shen2023fine}, HD-VILA-100M~\cite{xue2022advancing}, and AudioSet~\cite{gemmeke2017audio}. 
As detailed in Table~\ref{tab:dataset_stat}, MM-V2A aggregates 69,749 videos from VIDAL-10M, 44,637 videos from HD-VILA-100M, 128,079 videos from AudioSet, and the complete sets of VGG-Sound, VALOR-32K, BBC Sound Effects, and FAVDBench.

For the VAE model, we collect several dataset in various domain to achieve optimal reconstruction results, including general audio: AudioSet,  FSD50K\footnote{https://freesound.org/}, FAVDBench, UrbanSound8K~\cite{salamon2014dataset}, VGGSound, VIDAL-10M, HD-VILA-100M, sound effect: BBC Sound Effect Library, music: MTG-Jamendo~\cite{bogdanov2019mtg}, and human voice: Common Voice Corpus 1 (English)\footnote{https://commonvoice.mozilla.org/en/datasets}, results in total number of 996,033 audio files, we refer the above collection of datasets as the VAE ensemble dataset.

\subsection{Baseline}
We adopt two advanced V2A models as baselines: 1) SpecVQGAN \cite{iashin2021taming}, a transformer-based autoregressive model that generates spectrogram VQVAE indices from visual features; and 2) Diff-Foley \cite{luo2024diff}, a powerful V2A model based on latent diffusion. For SpecVQGAN, we evaluate its best-performing variant using Resnet+Flow as visual input conditions. For Diff-Foley, we assess its performance both with and without classifier guidance to analyze the influence of its sophisticated external alignment mechanism.

\subsection{Video-to-Audio generation results}
We present the quantitative results on VGGSound test set in Table~\ref{tab:v2a_results}. For both Tri-Ergon-S and Tri-Ergon-L, we only use the video embedding $\mathbf{e_V}$ for fair compression. 
As we can observe that, our proposed Tri-Ergon-S outperforms previous methods significantly on both audio quality and temporal synchronization. 
Benefited from our newly proposed dataset, Tri-Ergon-L further pushes the boundaries of video-to-audio generation performance.

Figure~\ref{fig:demo_lufs} presents the qualitative results for Tri-Ergon-S using $\mathbf{e_V}$, $\mathbf{e_L}$, and $\mathbf{E_{L}}$ as the composed conditions, as predicted by $T_{LUFS-S}$. 
The $T_{LUFS-S}$ successfully predicted the momentary LUFS change over time, guiding the model to generate synchronous audio. Additionally, the measured LUFS of the generated audio closely aligns with the input $\mathbf{E_{L}}$ condition, demonstrating the fine-grained controllability of the proposed Tri-Ergon.

\subsection{Fine-grained Condition Analysis}
\begin{figure}[t]
    \centering
    \includegraphics[width=0.8\linewidth]{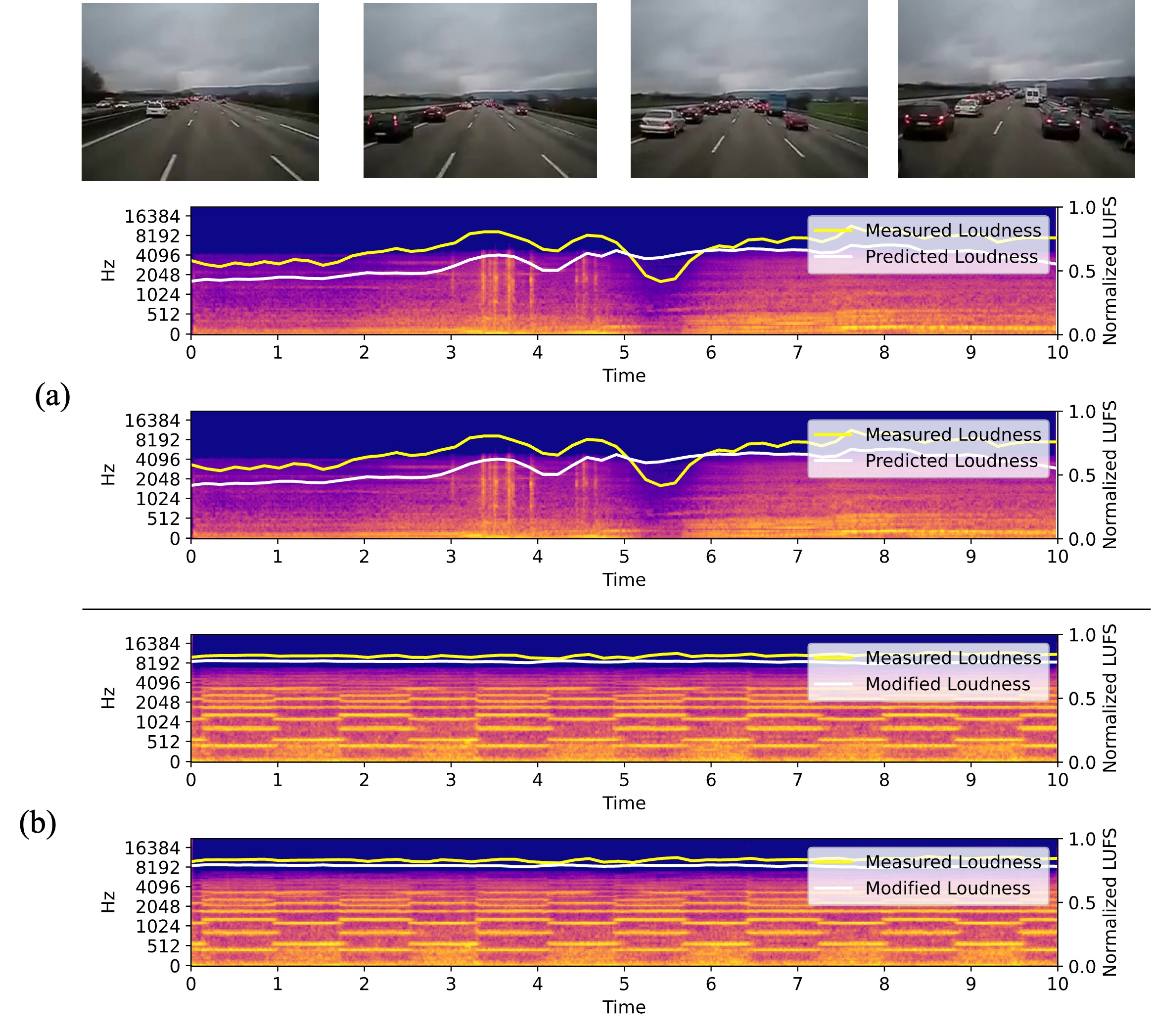}

    \caption{Fine-grained Condition Analysis. We present a scenario featuring an ambulance with no visible cue on a roadway. (a) Generated audio with visual information only. (b) Generated audio after adjusting the LUFS embedding along with additional text prompt.}

    \label{fig:condition}
\end{figure}

Our proposed model captures fine-grained multimodal information to enable more precise control. As illustrated in Figure~\ref{fig:condition}, the absence of an ambulance in the scene results in the model's LUFS prediction lacking any visual information about an ambulance, leading to the generation of traffic sounds on the street, as shown in Figure~\ref{fig:condition}(a). 
However, by manually adjusting the LUFS embedding and incorporating a text prompt specifying the ambulance sound \enquote{Sound of loud ambulance siren}, the model can generate audio that includes the sound of an ambulance, as shown in Figure~\ref{fig:condition}(b).

\subsection{Ablation Study}
\subsubsection{Compose Different Conditions.}
As discussed above, different condition types provide diverse guidance for the final generation results.
In this  section, we study the impact of different combinations of $\mathbf{e_V}$, $\mathbf{e_L}$ and $\mathbf{e_A}$ on the VGGSound dataset, as shown in Table~\ref{tab:s_vgg_ablation}.
We also compare the results between using $\mathbf{E_L}$ predicted by $T_{LUFS}$ and the ground truth of the testing audio, the latter can be seen as handcrafted momentary LUFS by audio engineers. We observe the following: 1) Additional text embedding will improve the audio semantic relevance with the video, but have marginal contribution to the audio-video synchronization. 2) Using the ground truth $\mathbf{e_A}$ significantly improves the semantic similarity, demonstrating the one-shot flexibility of Tri-Ergon in audio generation.
3) Utilizing ground truth $\mathbf{E_L}$ further enhances both audio quality and alignment, demonstrating the versatility and flexibility of our model, which allows for manual intervention when the results do not meet the user's requirements.
\subsubsection{CFG Scale \& Number of Inference Steps analysis.}
\begin{figure}[!ht]
    \centering
        \includegraphics[trim = 6mm 0 0 8mm, clip, width=0.49\columnwidth]{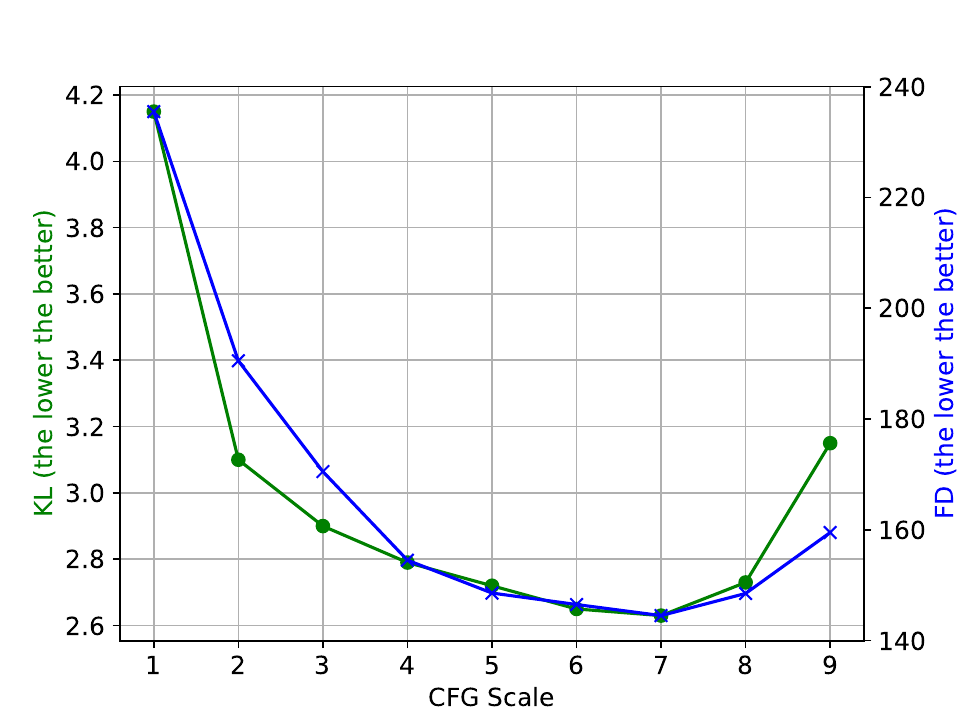}
        \includegraphics[trim = 6mm 0 0 8mm, clip,width=0.49\columnwidth]{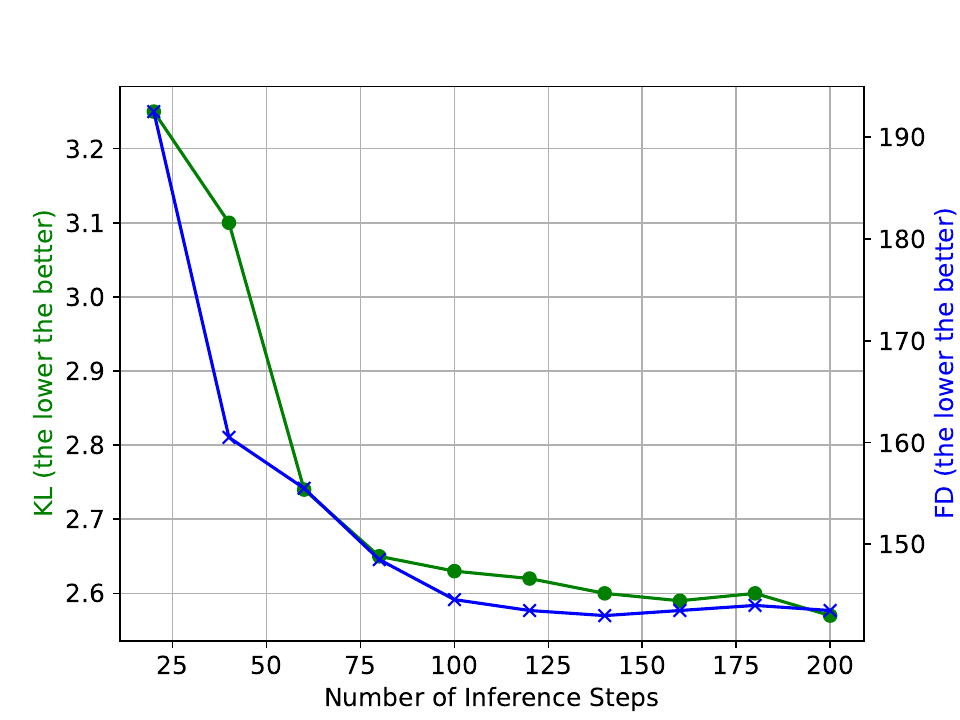}
        % \vspace{-3mm}
    \caption{CFG Scale \& Number of Inference Steps. Our model reach the best performance with $\omega=7$, inference steps of 100. Inference with more steps after 100 only slightly improve the audio quality.}
    \label{fig:KL_FD_comparison}
\end{figure}

As shown in Figure~\ref{fig:KL_FD_comparison}, our model achieves optimal performance with a CFG scale $\omega$ set to 7 and an inference step count of 100. Further increasing the number of inference steps beyond 100 results in only marginal improvements in audio quality. Therefore, for balancing computational efficiency and audio fidelity, 100 inference steps with a CFG scale of 7 is the recommended configuration.

\begin{table}[t]
\centering
\small
\setlength{\tabcolsep}{0.11cm}
\begin{tabular}{@{}lllll|lll@{}}
\toprule
$\mathbf{e_V}$ &$\mathbf{e_L}$ &$\mathbf{\overline{e}_A}$             & $\mathbf{\widehat{E}_L}$ & $\mathbf{\overline{E}_L}$       & $\text{FD}_{openl3}$ $\downarrow$ & $\text{KL}_{passt}$ $\downarrow$ & AV-Align $\uparrow$      \\ \midrule
\ding{52} & \ding{56} & \ding{56} & \ding{52}  & \ding{56} & $144.57$                           & $2.63$                             & $0.229$          \\
\ding{52} & \ding{52} & \ding{56} & \ding{52}  & \ding{56} & $141.47$                            & $2.61$                             & $0.229$          \\
\ding{52} & \ding{56} & \ding{52} & \ding{52}  & \ding{56} & $103.82$                             & $1.83$                             & $0.236$          \\
\ding{52} & \ding{52} & \ding{52} & \ding{52}  & \ding{56} & $96.23$                    & $1.47$                             & $0.240$          \\
\ding{52} & \ding{56} & \ding{56} & \ding{56}  & \ding{52} & $89.92$                            & $1.79$                             & $0.232$          \\
\ding{52} & \ding{52} & \ding{56} & \ding{56}  & \ding{52} & $89.65$                            & $1.608$                             & $0.234$          \\
\ding{52} & \ding{56} & \ding{52} & \ding{56}  & \ding{52} & $82.29$                             & $1.06$                             & $0.238$          \\
\ding{52} & \ding{52} & \ding{52} & \ding{56}  & \ding{52} & $\mathbf{78.19}$                             & $\mathbf{0.97}$                    & $\mathbf{0.242}$ \\ \bottomrule
\end{tabular}
\caption{\footnotesize{The effect of using different combinations of conditions with Tri-Ergon-S on VGGSound. 
We use $\mathbf{\overline{e_A}}$ denote ground truth $\mathbf{e}_A$, $\mathbf{\widehat{E}_L}$ denote $\mathbf{E_L}({T_{LUFS}})$, and $\mathbf{\overline{E}_L}$ denote $\mathbf{E_L}({GT})$.} \label{tab:s_vgg_ablation}}
\end{table}

\section{Conclusion}
\label{conclusion}

The Tri-Ergon model is a significant advancement in V2A synthesis, addressing key limitations by integrating multi-modal inputs and fine-grained loudness control. It incorporates textual, auditory, and visual prompts, achieving a richer, more nuanced audio generation aligned with the video context. LUFS embedding enhances its utility in professional settings, particularly in Foley workflows. It can produce high-fidelity stereo audio up to 60 seconds, surpassing the current state-of-the-art and opening new possibilities for detailed sound synthesis in various applications. These advancements have transformative potentials for film, gaming, virtual reality, and automated video processing industries. Moving forward, refining these techniques and exploring their applications in different contexts will be crucial, such as audible-video generation~\cite{mao2024tavgbench} or techniques for efficient generation~\cite{qin2024you}.

\bibliography{aaai25}

\end{document}